\pdfoutput=1

\documentclass[11pt]{article}

\usepackage{acl}

\usepackage{times}  
\usepackage{latexsym}

\usepackage[T1]{fontenc}

\usepackage[utf8]{inputenc}

\usepackage{microtype}

\usepackage{inconsolata}

\usepackage{amsmath}
\usepackage{makecell}
\usepackage{multirow}
\usepackage[cjk]{kotex}
\usepackage{array}
\usepackage{booktabs}
\usepackage{graphicx}
\usepackage{array}
\usepackage{colortbl}
\usepackage{amsfonts} 
\usepackage{tabularx}

\usepackage[framemethod=TikZ]{mdframed}
\usepackage{xcolor}

\definecolor{cadmiumgreen}{rgb}{0.0, 0.42, 0.24}

\title{\textsc{NeedleChain}: Measuring Intact Context Comprehension Capability of Large Language Models}

\author{Hyeonseok Moon, Heuiseok Lim$^{\dagger}$\\
Department of Computer Science and Engineering, Korea University\\
\texttt{\{glee889,limhseok\}@korea.ac.kr}
}

\begin{document}
\maketitle
\begin{abstract}

Recent reports suggest that LLMs can handle increasingly long contexts. However, many existing benchmarks for context understanding embed substantial query-irrelevant content, which shifts evaluation toward retrieving relevant snippets rather than fully integrating all provided information. Under this setting, we view that current benchmarks can overestimate true context-understanding ability of LLMs. In particular, we demonstrate that when the context consists entirely of query-relevant text, even advanced models such as GPT-4o fail to reliably integrate inputs as short as 200 tokens. To evaluate this capability more rigorously, we introduce \textsc{NeedleChain}, a benchmark designed to test whether models can faithfully incorporate all given evidence. \textsc{NeedleChain} includes three variants that differ in the required order of comprehension, along with a parallel benchmark based on the needle-in-a-haystack(NIAH) paradigm. By comparing these variants, \textsc{NeedleChain} enables a more comprehensive assessment of context understanding. We further propose a training-free strategy that encourages models to reflect all available information, ROPE contraction, highlighting the importance of full-context integration and pointing to new directions for improving reliable reasoning over context.

\end{abstract}

\section{Introduction}

Over the past few years, we have witnessed significant advancement of large language models (LLMs) in various aspects \cite{yang2025qwen3, hurst2024gpt, team2025gemma}. Enhanced reasoning abilities now allow LLMs to tackle several challenges and expand the range of tasks they can address \cite{liu2024deepseek, team2025qwq}. In particular, recent work has substantially extended the maximum context length of LLMs, enabling them to handle more complex tasks \cite{tworkowski2023focused, ge2025a, jin2024llm, ding2024longrope}. For example, Llama-2 \cite{touvron2023llama} supports a 4,096-token context window, whereas Llama-4 \cite{meta2025llama} reportedly scales to over one million tokens. These advances suggest markedly improved ability to model and utilize long-range dependencies in the input.

\begin{figure}[t]
\centering
    \includegraphics[width=\linewidth]{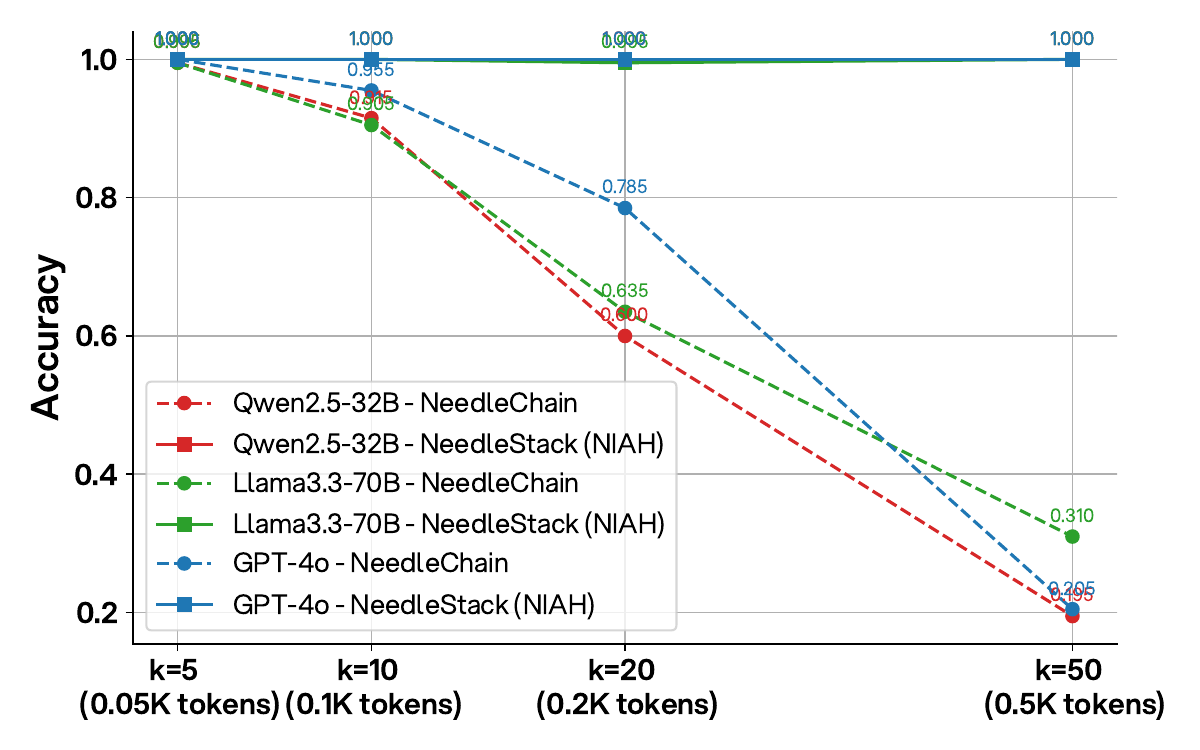}
    \caption{
    Performance comparison between the \textsc{NeedleChain} (Backward chain) and its parallel NIAH paradigm benchmark (NeedleStack). Reported number of tokens were estimated with Qwen2.5 tokenizer.
    }
    \label{fig:niah}
\end{figure}



\begin{figure*}[t]
\centering
\includegraphics[width=0.9\linewidth]{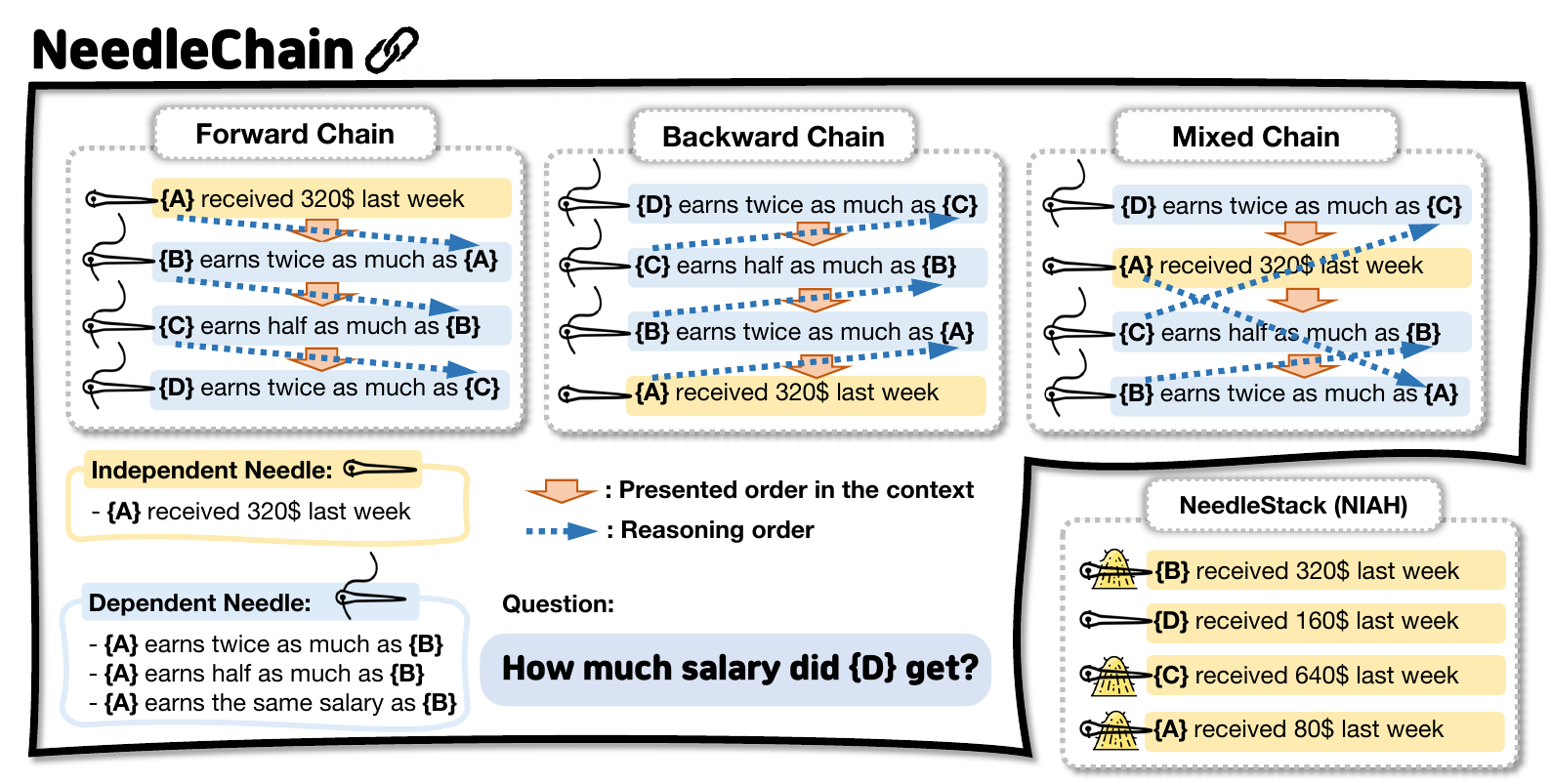}
 \caption{Performance variation with respect to the domain composition of training data}
 \label{fig:overall}
\end{figure*}

However, we question whether LLMs have reached to the point where they can reliably recognize and interpret all provided context. Numerous long-context benchmarks evaluate LLMs by presenting a lengthy passage that intermixes query-relevant and query-irrelevant information \cite{bai-etal-2024-longbench}. The needle-in-a-haystack (NIAH) paradigm pushes this to an extreme by filling the context with misleading content while providing only a small amount of query-relevant evidence \cite{li2024needlebench, schuster2025needle, NEURIPS2024_24a8968a}. We argue that such settings can overstate contextual understanding: evaluation might prioritize \textit{retrieval of the necessary snippet} rather than verifying whether the model tracks and integrates \textit{all} information in the context.

In this paper, we cast a simple yet overarching question: \textit{"When the context contains only query-relevant information, can an LLM fully incorporate and understand it?"} Prior studies have notified a large gap between the reported context length and the effective length models can use, under the NIAH paradigm \cite{kuratov2024babilong, hsieh2024ruler}. We go further by characterizing the intact context-understanding length, i.e., the maximum context length over which the model preserves the ability to reflect the entire context without omissions.


In particular, we introduce \textbf{\textsc{NeedleChain}}, a benchmark that rigorously evaluates holistic context comprehension. 
In \textsc{NeedleChain}, each statement forms part of a causal chain and remains necessary for deriving the correct answer; omitting any single element prevents a correct solution. We build the benchmark from short, tightly connected statements (e.g., \textit{"{A} received \$1,600 last week"} and \textit{"{A} earns twice as much as {B},"}) that the model must jointly integrate to answer a query. Specifically, we formalize \textbf{reasoning order}, which denotes the logical sequence in which a model must process contextual information to resolve a query. We instantiate three variants: \emph{forward chain} (requires left-to-right integration), \emph{backward chain} (requires right-to-left integration), and \emph{mixed chain}. Beyond these variants, we construct an NIAH dataset, \textbf{NeedleStack}, from the same benchmark component (needles). Figure~\ref{fig:overall} illustrates representative examples, and subsequent sections describe the construction process in detail. Comparing these variants exposes previously overlooked weaknesses in LLMs’ ability to preserve and integrate all the context intactly. Notably, figure~\ref{fig:niah} contrasts the performance on \textsc{NeedleChain} (backward) benchmark with the NeedleStack (NIAH) benchmark. Our findings reveal that even with a context of just 200 tokens, LLMs often fail to reliably retain and combine all information required to answer the query.

Through detailed error analysis, we identify factors that hinder faithful context understanding. 
We further propose a ROPE contraction strategy as a simple yet effective method to enhance context comprehension. This strategy sets a larger ROPE \cite{su2024roformer} embedding rotation angle at inference than at training time, thereby sharpening positional distinctions and strengthening context utilization. Overall, we argue that beyond simply extending a model’s context length, its ability to fully utilize the provided context constitutes a critical factor.


\section{\textsc{NeedleChain}}

\textsc{Needlechain} is designed as an information-dense context comprehension task. In each context, every sentence contains essential content, requiring the model to fully understand all details for successful completion. Below, we discuss the details related to data construction.

\subsection{Needle Design}
We design our benchmark to analyze the weaknesses of LLMs in processing context by treating the entire context as a single semantic unit and defining a reasoning order required to understand the information provided in the context. To achieve this, we introduce the concept of a "needle" to create connections between preceding and succeeding information. We define two types of needles:

\begin{itemize}
    \item \textbf{Independent Needle} This refers to a sentence that contains independent information without relying on other context. (\textit{"\{A\} received \$1600 last week"})
    \item \textbf{Dependent Needle} This refers to a sentences that provides information in conjunction with other sentences. It is designed to create an extended context while maintaining a coherent semantic unit. (\textit{"\{A\} earns twice as much as \{B\}"} / \textit{"\{A\} earns half as much as \{B\}"} / \textit{"\{A\} earns the same salary as \{B\}"})
    \end{itemize}

In particular, we design three types of dependent chains—"halving," "doubling," and "retaining"—that require clear yet straightforward reasoning. This design allows us to evaluate the reasoning capabilities of LLMs without underestimating their core language comprehension abilities due to their lack of precision in complex mathematical calculations, such as decimal operations. We combine these needles according to the specific objectives of our benchmark to form the final evaluation suites.

\subsection{Chain Composition}

Each data point (chain) of \textbf{NeedleChain} consists of one independent needle and $k-1$ dependent needles. By leveraging these needles, the combined chain is designed to maintain a single sequential reasoning order (e.g. \textit{A is related to B, B is related to C, C is related to D}). We then design a question about the last needle in the reasoning order to ensure all context is relevant to the query (\textit{How much salary did \{D\} get?}). Specifically, we propose three variants as follows. Figure~\ref{fig:overall} illustrates examples for each chain composition.

\begin{itemize}
    \item {\textbf{Forward Chain}} The reasoning process must proceed in a \textbf{left-to-right} sequence. Accurate conclusions can only be drawn by following the presented order of the given input.
    \item {\textbf{Backward Chain}} The reasoning process must follow a \textbf{right-to-left} sequence. LLM must track the contextual information in the reversely-presented order, starting from the most recently presented data.
    \item {\textbf{Mixed Chain}} The sequence of required reasoning steps is set \textbf{arbitrarily}. The LLM must identify the randomly given reasoning order to arrive at the correct answer.
\end{itemize}

We design three chain variants using identical needle composition, distinguished solely by the sequence of provided information, thereby ensuring the same reasoning order while different presented order across chain variants. We determine our benchmark's evaluation suite by posing questions related to the needle at the end of reasoning order. This approach ensures our benchmark is structured so that answering the questions correctly requires 1) following the designated path and 2) fully understanding the given context. Comparing the performance of these variants enables clearly characterizing the limitations of LLMs in contextual understanding.

Furthermore, we use our needles to construct data under the NIAH paradigm, which concatenates multiple independent needles into a single context. In this setting, when we ask a question about one needle, all remaining needles serve as irrelevant context (i.e., the haystack). We refer to this benchmark as "NeedleStack (NS)" By comparing these variants of \textsc{NeedleChain}, we effectively demonstrate context-comprehension capabilities of LLMs and clearly distinguish performance on sparse contexts versus information-dense contexts.

\definecolor{metacolorab}{HTML}{F0F0F0}

\begin{table*}[t]

\centering
\resizebox{0.9\linewidth}{!}{
\footnotesize
\begin{tabular}{l|cccc|cccc|cccc}

\toprule[1.5pt]

\multicolumn{1}{c|}{\multirow{3}{*}{\textbf{Model}}} & 
\multicolumn{4}{c|}{\quad k=5 (Token Length: 0.05K) \quad} &
\multicolumn{4}{c|}{\quad k=10 (Token Length: 0.1K) \quad} &
\multicolumn{4}{c}{\quad k=20 (Token Length: 0.2K) \quad} \\

{} &
\multirow{2}{*}{\textbf{NS}} & \multicolumn{3}{c|}{\textbf{NeedleChain}} & 
\multirow{2}{*}{\textbf{NS}} & \multicolumn{3}{c|}{\textbf{NeedleChain}} & 
\multirow{2}{*}{\textbf{NS}} & \multicolumn{3}{c}{\textbf{NeedleChain}} \\

{} &
& \textbf{F} & \textbf{B} & \textbf{M} &
& \textbf{F} & \textbf{B} & \textbf{M} &
& \textbf{F} & \textbf{B} & \textbf{M} \\
\cmidrule(lr){1-1}\cmidrule(lr){2-2}\cmidrule(lr){3-3}\cmidrule(lr){4-4}\cmidrule(lr){5-5}
\cmidrule(lr){6-6}\cmidrule(lr){7-7}\cmidrule(lr){8-8}\cmidrule(lr){9-9}
\cmidrule(lr){10-10}\cmidrule(lr){11-11}\cmidrule(lr){12-12}\cmidrule(lr){13-13}

\textbf{Qwen3-32B} & 100.0 & 100.0 & 99.5 & 100.0 & 100.0 & 99.5 & 91.5 & 98.0 & 99.5 & 94.0 & 65.0 & 86.0 \\
\textbf{Qwen2.5-32B} & 100.0 & 100.0 & 99.5 & 99.5 & 100.0 & 98.0 & 91.5 & 96.5 & 100.0 & 95.0 & 60.0 & 89.5 \\
\textbf{QwenLong-L1} & 100.0 & 100.0 & 100.0 & 100.0 & 100.0 & 100.0 & 100.0 & 100.0 & 100.0 & 98.5 & 92.5 & 99.5 \\
\textbf{Llama3.3-70B} & 100.0 & 100.0 & 99.5 & 99.5 & 100.0 & 100.0 & 90.5 & 98.5 & 99.5 & 94.0 & 63.5 & 91.0 \\
\textbf{GPT-4o} & 100.0 & 100.0 & 100.0 & 99.5 & 100.0 & 100.0 & 95.5 & 98.5 & 100.0 & 98.0 & 78.5 & 88.5 \\

\midrule[1.5pt]
\end{tabular}
}

\resizebox{0.92\linewidth}{!}{
\footnotesize
\begin{tabular}{l|cccc|cccc|cccc}


\multicolumn{1}{c|}{\multirow{3}{*}{\textbf{Model}}} & 
\multicolumn{4}{c|}{\quad k=50 (Token Length: 0.5K) \quad} &
\multicolumn{4}{c|}{\quad k=100 (Token Length: 1K) \quad} &
\multicolumn{4}{c}{\quad k=200 (Token Length: 2K) \quad} \\

{} &
\multirow{2}{*}{\textbf{NS}} & \multicolumn{3}{c|}{\textbf{NeedleChain}} & 
\multirow{2}{*}{\textbf{NS}} & \multicolumn{3}{c|}{\textbf{NeedleChain}} & 
\multirow{2}{*}{\textbf{NS}} & \multicolumn{3}{c}{\textbf{NeedleChain}} \\

{} &
& \textbf{F} & \textbf{B} & \textbf{M} &
& \textbf{F} & \textbf{B} & \textbf{M} &
& \textbf{F} & \textbf{B} & \textbf{M} \\
\cmidrule(lr){1-1}\cmidrule(lr){2-2}\cmidrule(lr){3-3}\cmidrule(lr){4-4}\cmidrule(lr){5-5}
\cmidrule(lr){6-6}\cmidrule(lr){7-7}\cmidrule(lr){8-8}\cmidrule(lr){9-9}
\cmidrule(lr){10-10}\cmidrule(lr){11-11}\cmidrule(lr){12-12}\cmidrule(lr){13-13}

\textbf{Qwen3-32B} & 99.5 & 80.5 & 13.5 & 46.5 & 99.5 & 68.5 & 10.5 & 10.5 & 99.5 & 47.5 & 1.5 & 4.5 \\

\textbf{Qwen2.5-32B} & 100.0 & 81.0 & 19.5 & 44.5 & 100.0 & 65.5 & 7.0 & 16.5 & 100.0 & 43.0 & 0.5 & 8.5 \\
\textbf{QwenLong-L1} & 100.0 & 88.5 & 46.0 & 76.0 & 99.5 & 86.5 & 21.5 & 39.0 & 99.5 & 67.8 & 6.5 & 11.0 \\
\textbf{Llama3.3-70B} & 100.0 & 76.5 & 31.0 & 79.0 & 100.0 & 67.5 & 22.5 & 45.0 & 99.5 & 44.0 & 18.0 & 12.5 \\
\textbf{GPT-4o} & 100.0 & 76.5 & 20.5 & 61.0 & 100.0 & 36.0 & 7.0 & 26.0 & 98.0 & 14.0 & 4.0 & 5.0 \\

\bottomrule[1.5pt]
\end{tabular}
}

\caption{Performance of several LLMs on NeedleChain (\textbf{NS}: NeedleStack, \textbf{F}: Forward chain, \textbf{B}: Backward, \textbf{M}: Mixed Chain).
}

\label{tb:main}
\end{table*}

\subsection{Benchmark Details}

In constructing the needle, we utilized a randomly selected name list officially released by the U.S. government \footnote{\url{https://www.ssa.gov/oact/babynames/decades/century.html}}. For each data point, we first established a name list, then used this list to create three chain variants and a NeedleStack. This approach allowed us to apply the same name list across our benchmark variants, thereby minimizing unintended bias related to naming \cite{eloundou2025firstperson}.

As our benchmark uses synthetic data, we can adjust the context length by varying the number of needles. Under this setup, we investigate context comprehension in LLMs by examining performance as the total number of needles, denoted by $k$, increases up to 200. While we theoretically could increase $k$ further, even relatively short contexts (approximately 2K tokens) already revealed clear trends in the models’ context understanding capacity that we aim to analyze. Thus, we conducted experiments using only 5 to 200 needles. We generated 200 test instances for each dataset.

\section{Experiments}

We conduct experiments using state-of-the-art LLMs reported to possess long context understanding capabilities. In particular, we focus our experiments on the widely used LLMs: Qwen2.5-32B \cite{qwen2.5}, QwenLong-L1 \cite{qwen2.5}, Qwen3-32B \cite{yang2025qwen3}, Llama3.3-70B \cite{grattafiori2024llama}, and GPT-4o \cite{hurst2024gpt}. Detailed information about the models, prompts used for evaluation, and the evaluation environment is provided in the Appendix~\ref{app:evaluation_details}.

\subsection{Main Results}
We first evaluate the performance of current LLMs on our \textsc{NeedleChain} and NeedleStack benchmarks. The experimental results are presented in Table~\ref{tb:main}. Key insights from this study are as follows:

\paragraph{Limitations in Long-Context Understanding} 
All LLMs achieve near-perfect performance on NeedleStack (the NIAH paradigm). However, accuracy on our \textsc{NeedleChain} benchmark begins to drop once $k$ exceeds 10 and cannot remain high when $k$ reaches 50. This pattern suggests that models still fail to fully exploit context even when the total sequence length is only 0.5K tokens. Although the LLMs we evaluate nominally support much longer context windows (GPT-4o: 16K, Qwen2.5-32B: 32K, Qwen3-32B: 32K, QwenLong-L1: 1M, Llama3.3-70B: 128K), NeedleChain exposes clear limitations in their practical context understanding that existing benchmarks do not reveal.

\paragraph{LLMs Struggle to find solution on Backward Chain}
We observed a particularly pronounced performance decline in the backward chain. This finding indicates that the LLM's ability to use context depends strongly on the reasoning direction, revealing a clear vulnerability in reverse reasoning. The backward chain even showed a larger performance drop than mixed chains with arbitrary reasoning paths. The higher performance in mixed chains suggests that text which merely \textit{appears complex} may not substantially challenge LLMs. Instead, the main difficulty lies not in the seemingly intricate reasoning paths, but in the requirement to reason in the reverse direction itself.

\paragraph{Forward-direction reasoning is suitable for LLMs}
Among the three chain variants in the \textsc{NeedleChain} benchmark, comprehension in the forward direction is notably high. This suggests that LLMs can comfortably comprehend the given context when its logical order aligns with their left-to-right processing. This observation not only reveals limitations of current LLMs, but indicates that we can maximize their reasoning capabilities by presenting information sequentially.

\begin{figure*}[ht]
\centering
\includegraphics[width=0.95\linewidth]{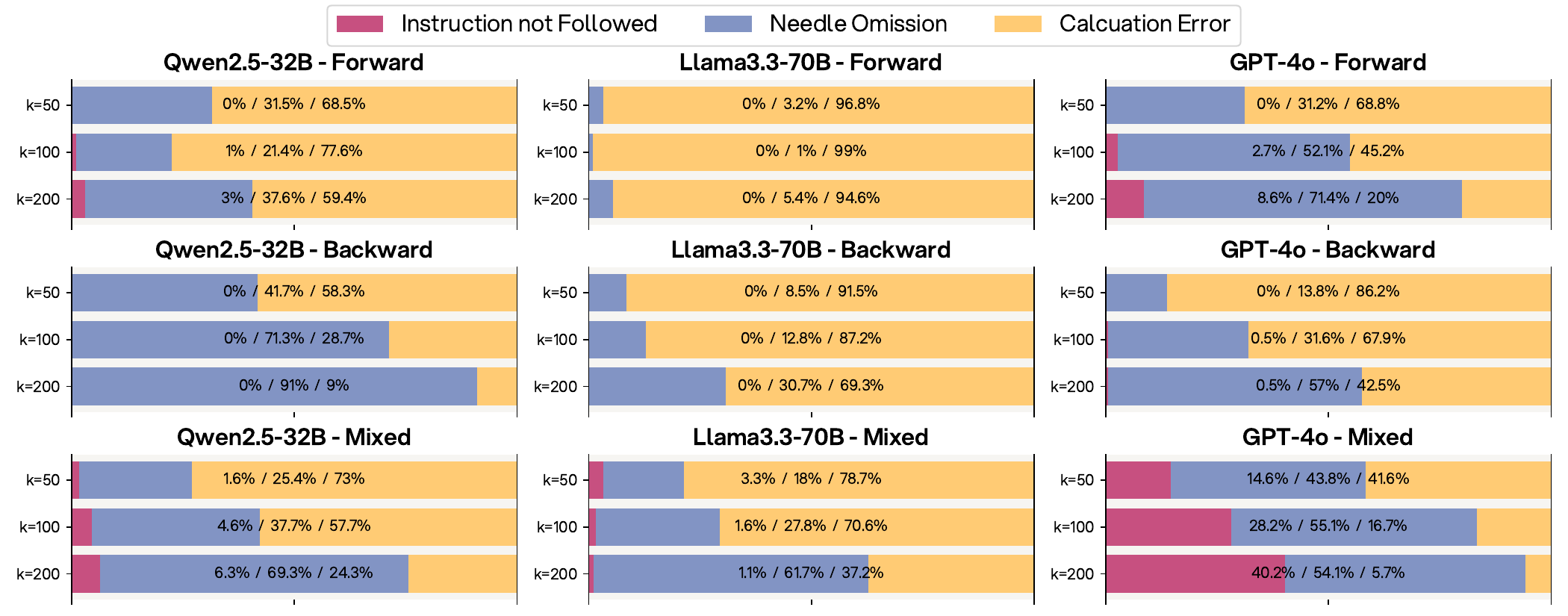}
 \caption{Error analysis on \textsc{NeedleChain}. We analyze errors in each category to determine which of the three predefined error types they fall into.}
 \label{fig:analysis}
\end{figure*}

\subsection{Error Analysis}
\label{sec:error_analysis}
To gain a deeper understanding of the weaknesses in LLMs' context comprehension as identified by the \textsc{NeedleChain} benchmark, we analyze the error cases revealed by our benchmark. We categorize the errors in LLMs into three distinct types and find that this taxonomy successfully encompasses all observed error cases:
\begin{itemize}
    \item \textbf{Instruction not Followed}:  This refers to instances where the model fails to generate a response by not adhering to the given output format, or fails to determine the final answer.
    \item \textbf{Needle Omission} This refers to cases when certain "needles" are omitted in generating final answer. Specifically, it refers to cases where the name provided as input is absent in the output, resulting in an incorrect answer.
    \item \textbf{Calculation Error}  This pertains to situations where the intermediate steps are correct, but an error occurs in the final answer computation. If an error does not fall into the first two categories, we classify such case here.
\end{itemize}

Concrete examples for each error type are contained in the Appendix~\ref{app:qualitative}. Using this taxonomy, we analyze error cases made by LLMs in our benchmark. Figure~\ref{fig:analysis} displays the results. Key takeaways from our analysis are summarized as follows:

\paragraph{For small k:} 
We find that calculation errors are the dominant source of error, particularly when k is small (k=50, in this experiment). We can observe this phenomenon consistently across all models, regardless of the reasoning order. Considering the substantial forward–backward performance gap in previous experiments (Table~\ref{tb:main}), this result further underscores the strong influence of reasoning order on calculation capability. For Llama3.3-70B, for instance, the forward–backward performance gap exceeds 40 percentage points at k=50, and the results in Figure~\ref{fig:analysis} indicate that this gap stems entirely from reduced calculation capability induced by the change in reasoning order.

\paragraph{For larger k:}
As we increase k, needle omission emerges as a primary source of error, particularly in the Mixed chain. Across all models, the proportion of errors due to needle omission increases approximately linearly with k. This implies that the model fails to incorporate a clearly identifiable portion of the given context, and we observe that this failure mode already appears frequently once the input length exceeds only 1K tokens (k=100). This pattern suggests that achieving robust context understanding remains an open challenge. We provide a more detailed analysis of this error type in a subsequent section.

\subsection{Position Heatmap}
In this section, we utilize \textsc{NeedleChain} to identify positional weaknesses of LLMs in the context comprehension. Note that our input consists of multiple needles, each carrying key information about a specific "name." Considering the final answer can only be derived when the information from all needles is reflected, if any name is omitted from the LLM's response, it indicates that the corresponding needle was not considered during generation.

\begin{figure*}[t]
\centering
\includegraphics[width=1\linewidth]{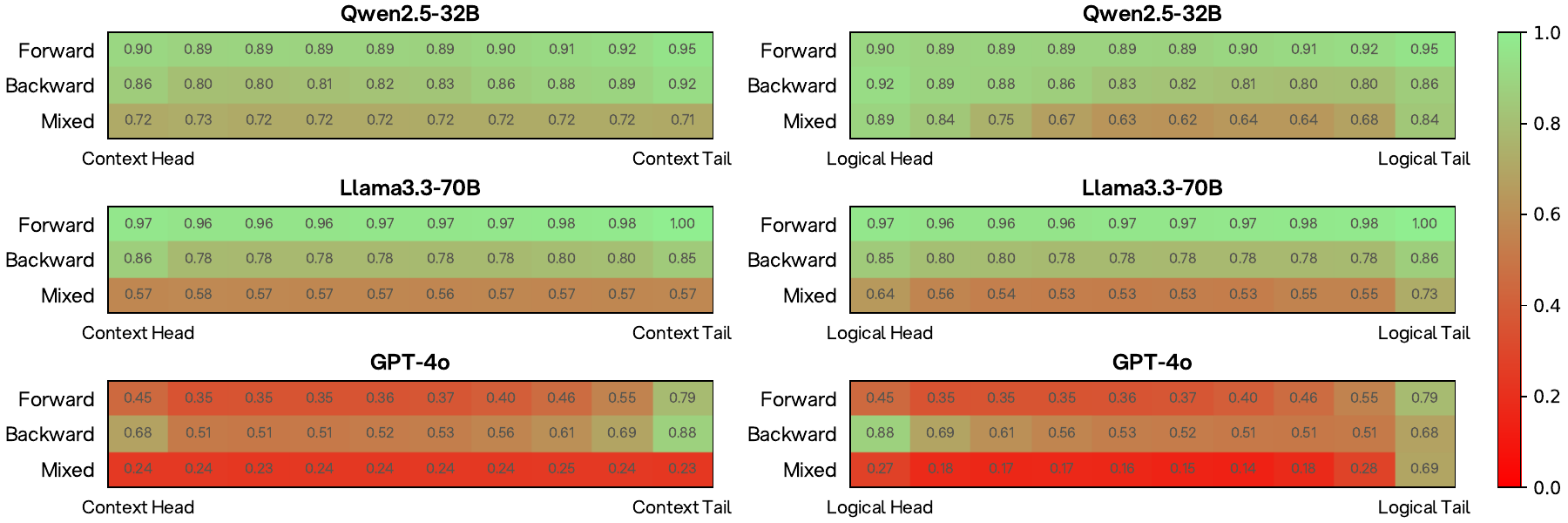}
 \caption{Heatmap to show the weaknesses for each position. Left-sided figures shows positional needle-missing heatmap with respect to the ``\textbf{presented order}". Right-sided figures shows those of ``\textbf{reasoning order}". We conducted experiments with k=200.}
 \label{fig:heatmap}
\end{figure*}

Through this approach, we identify positional weaknesses by determining whether the names given in the input are included in the generated text. We specifically analyze these positional weaknesses from two perspectives: presented order and reasoning order. We represent the ratio of names mentioned at each position relative to the total number of responses. The experimental results are shown in Figure~\ref{fig:heatmap}.

The results show that LLMs struggle to fully capture information from the input, as reflected by the unstable pattern of red regions in the heatmap, which highlights the \textbf{``logically lost-in-the-middle''} phenomenon \cite{liu2024lost}. We observe this most clearly in the mixed-chain setting. When we evaluate positions by their presented order in the context, we see little position-specific weakness: performance decreases fairly uniformly across all positions. In contrast, when we evaluate positions by reasoning order, the model’s ability to exploit information drops sharply at the "middle position". This suggests a practical takeaway: LLMs tend to \textit{get lost in the middle of the logical flow, rather than in the middle of the given context.}

\subsection{Case Study: Question Variants}
\begin{figure*}[ht]
\centering
\includegraphics[width=1\linewidth]{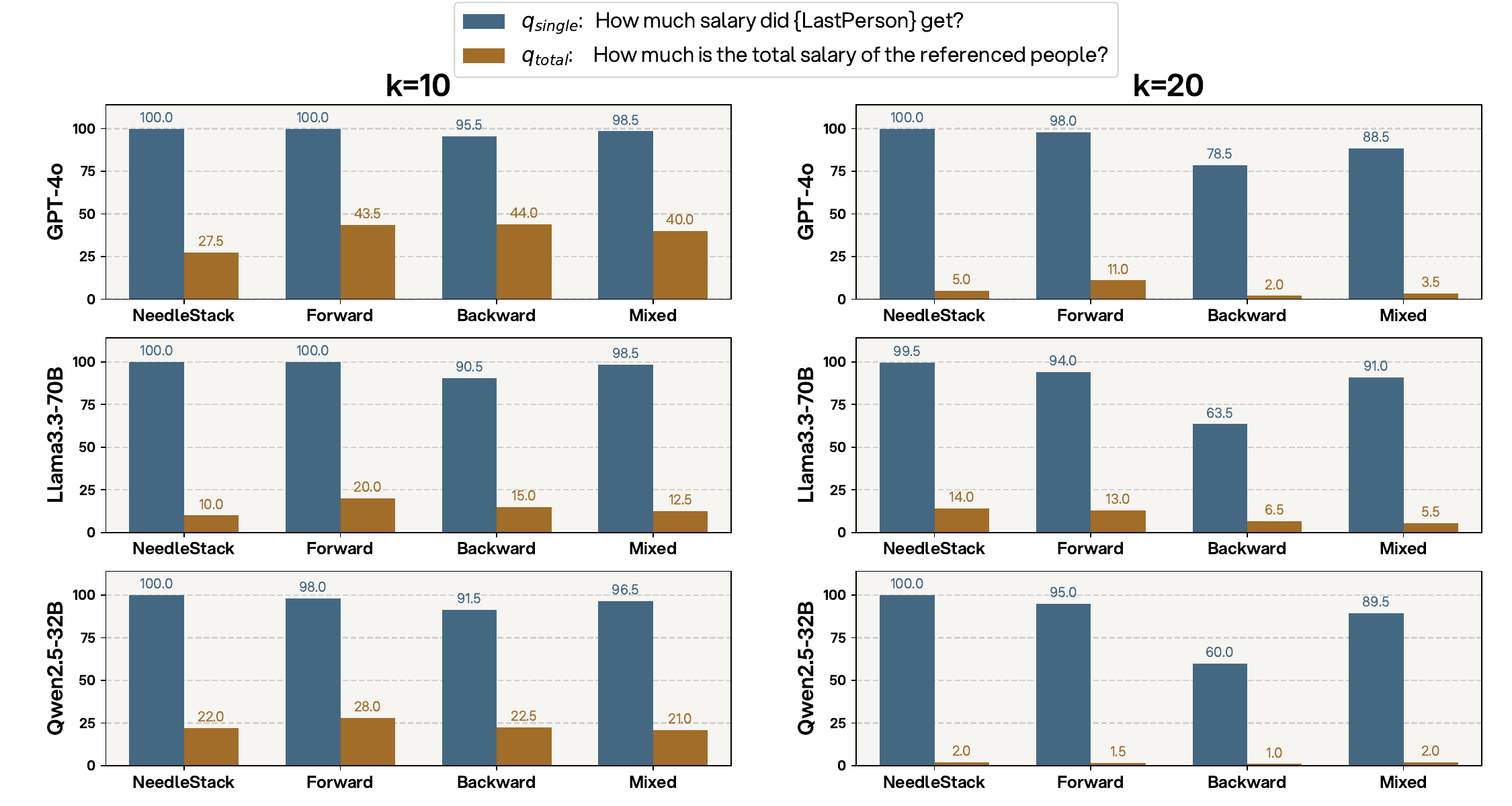}
 \caption{We compare the accuracy of models for different types of questions: those requires understanding the tail of the reasoning chain ($q_{single}$) and those requiring comprehensive understanding of the entire context ($q_{total}$).}
 \label{fig:question}
\end{figure*}

One can, in principle, construct an evaluation that requires aggregating all information in the given context simply by adjusting the query.  
For instance, even when using NeedleStack(NS), a question like "What is the total salary of all mentioned individuals?" requires the language model to integrate all context information.  
Building on this intuition, we propose two types of questions to more deeply analyze context understanding and to distinguish performance across these settings. Figure~\ref{fig:question} reports the experimental results.

The experimental results demonstrate that when using $q_{total}$, the performance of NS significantly declines, with scores dropping below 10 for all LLMs at $k = 20$. Notably, when using $q_{total}$, the performance on the \textsc{NeedleChain} benchmark exceeds that of the NS benchmark. This indicates that \textbf{deeper semantic connections (\emph{i.e.}, explicit reasoning paths within the context) enhance comprehension more than less interrelated contexts}. In a subsequent section, we further show how the design of \textsc{NeedleChain} provides additional insight into the context comprehension abilities of LLMs.

\subsection{Case Study: Tool Incorporation}
In Figure~\ref{fig:analysis}, we identified calculation errors as one of the critical factors contributing to the performance degradation of our benchmark. To address such error, recent approaches have sought to mitigate the mathematical limitations of LLMs by incorporating code implementation capabilities \cite{liao-etal-2024-mario, sharma-etal-2025-geocoder}. In this section, we aim to test whether computational deficiencies cause us to underestimate limitations in language comprehension by evaluating performance after code integration. Experimental results are detailed in Table~\ref{tb:tool}.

\definecolor{metacolorab}{HTML}{F0F0F0}


\begin{table}[ht]
\centering
\resizebox{1.0\linewidth}{!}{
\footnotesize
\begin{tabular}{ccr|l}

\toprule[1.5pt]
\multicolumn{1}{c}{\textbf{Type}} & 
\multicolumn{1}{c}{\textbf{Question}} & 
\multicolumn{1}{c|}{\textbf{k}} & 
\multicolumn{1}{c}{\textbf{Code Merging}} \\ 
\midrule[1.5pt]
\multicolumn{4}{c}{\textbf{NeedleStack}} \\
\midrule[1.5pt]
\textbf{NIAH} & $q_{total}$ & 5 & \textbf{96.0} $\rightarrow$ 94.0 ( \textcolor{blue}{$\nabla$ -2.0} ) \\
\textbf{NIAH} & $q_{total}$ & 10 & \textbf{27.5} $\rightarrow$ 53.0 ( \textcolor{red}{$\Delta$ +25.5} ) \\
\textbf{NIAH} & $q_{total}$ & 20 & \textbf{5.0} \,\,\,$\rightarrow$ 73.0 ( \textcolor{red}{$\Delta$ +68.0} ) \\
\textbf{NIAH} & $q_{total}$ & 50 & \textbf{6.5} \,\,\,$\rightarrow$ 72.0 ( \textcolor{red}{$\Delta$ +65.5} ) \\
\midrule[1.5pt]
\multicolumn{4}{c}{\textbf{NeedleChain}} \\
\midrule[1.5pt]

\textbf{Forward} & $q_{single}$ & 50 & \textbf{76.5} $\rightarrow$ 74.0 ( \textcolor{blue}{$\nabla$ -2.5} ) \\
\textbf{Forward} & $q_{single}$ & 100 & 36.0 $\rightarrow$ \textbf{42.5} ( \textcolor{red}{$\Delta$ +6.5} ) \\ \midrule
\textbf{Backward} & $q_{single}$ & 50 & 20.5 $\rightarrow$ \textbf{23.5} ( \textcolor{red}{$\Delta$ +3.0} ) \\
\textbf{Backward} & $q_{single}$ & 100 & \textbf{7.0} \,\,\,$\rightarrow$ 5.0 \,\,\,( \textcolor{blue}{$\nabla$ -2.0} ) \\ \midrule
\textbf{Mixed} & $q_{single}$ & 50 & \textbf{61.0} $\rightarrow$ 54.0 ( \textcolor{blue}{$\nabla$ -7.0} ) \\
\textbf{Mixed} & $q_{single}$ & 100 & \textbf{26.0} $\rightarrow$ 21.5 ( \textcolor{blue}{$\nabla$ -4.5} ) \\


\bottomrule[1.5pt]
\end{tabular}
}
\caption{Performance variation on tool merging. We report performance of GPT-4o model.
}
\label{tb:tool}
\end{table}


Interestingly, tool incorporation proved to be particularly effective in NeedleStack with $q_{total}$, where contextual information between segments is weakly correlated. The improvement from an apparent lack of understanding to performance above 70 points indicates that tool incorporation can be partially beneficial.

However, such incorporation did not prove effective in \textsc{NeedleChain}. We observed performance declines in most cases. This underscores the robustness of our benchmark, demonstrating that it presents challenges that can only be resolved with comprehensive long-context reasoning abilities. This finding suggests that the low performance of LLMs in \textsc{NeedleChain} is not solely due to computational limitations but rather reflects a deficiency in effectively integrating contextual information.

\section{Discussion}

\begin{figure}[ht]
\centering
    \includegraphics[width=\linewidth]{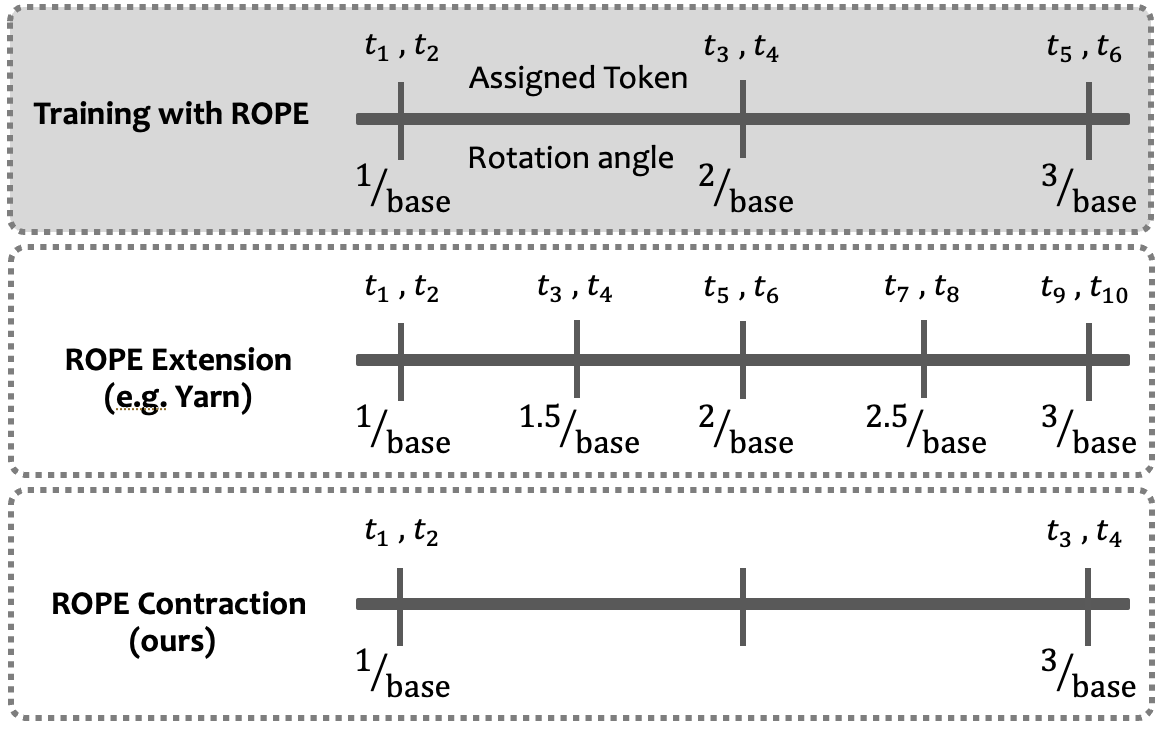}
    \caption{
    A simple diagram illustrating the concepts of ROPE extension and contraction
    }
    \label{fig:position}
\end{figure}

\begin{figure*}[t]
\centering
\includegraphics[width=1\linewidth]{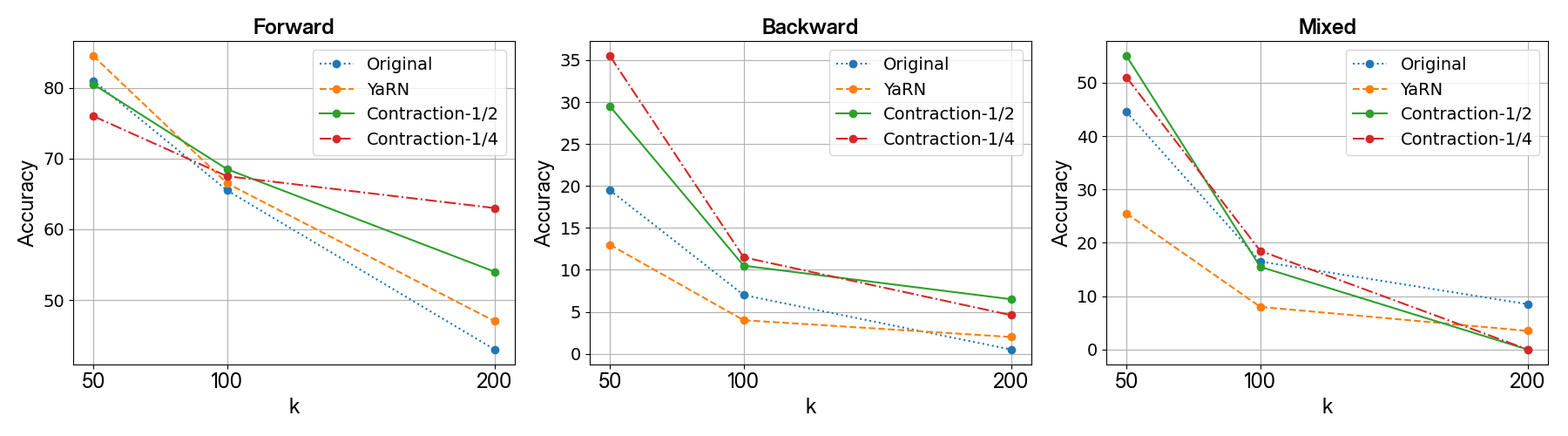}
 \caption{Performance variation derived by the ROPE contraction and extension methodologies}
 \label{fig:yarn}
\end{figure*}

Based on these discussions, we conclude that LLMs do not yet fully comprehend given contexts. \textbf{We argue that rather than hastily increasing the extent of context length, it might be beneficial to enhance comprehension within a limited range.} Accordingly, we revisit one of the strategies currently used to increase context length, namely ROPE extension \cite{zhong2025understanding, ding2024longrope, chen2023extending}. This approach modifies the rotary positional embeddings to theoretically extend the maximum usable context length. A common instantiation reduces the rotation angle used at inference time relative to that used during training. However, we hypothesize that such strategies might saturate positional distinctions and thereby substantially degrade the model’s ability to intactly utilize information across the entire context.

Conversely, we also hypothesize that sharpening positional distinctions can mitigate omissions of information within the given context, and we therefore propose ROPE contraction, a concept opposite to ROPE extension. A simple example is presented in Figure~\ref{fig:position}. For a simple toy experiment, we apply rotation angles that are 2x and 4x larger than those used during training to emphasize positional distinction.

We compare our results to the commonly used ROPE extension method known as Yarn \cite{peng2024yarn}. The results of these experiments are shown in Figure~\ref{fig:yarn}. The experimental results clearly support our argument. The use of Yarn significantly decreases performance on the \textsc{NeedleChain} benchmark, while the contraction method substantially improves the model’s ability to use the full context faithfully. This paper demonstrates that even simple methodologies can lead to meaningful performance improvements, paving the way for future research in this area.

The experimental results clearly support our argument. The use of Yarn significantly decreases performance on the NeedleChain benchmark, while the contraction method substantially improves the ability to understand context intactly. This paper demonstrates that even simple methodologies can lead to meaningful performance improvements, paving the way for future research in this area.

\section{Related Works}

There are currently numerous approaches to objectively assess the context understanding capabilities of LLMs \cite{bai-etal-2024-longbench, hsieh2024ruler, kuratov2024babilong, an-etal-2024-l, li-etal-2024-loogle, zhang-etal-2024-bench}. Among these, the Needle-in-a-Haystack benchmark is a widely used tool to evaluate long-range context understanding in LLMs \cite{yu2025sequential, song-etal-2025-counting}. However, this benchmark typically uses contexts in which most of the information is irrelevant to the query, which limits its ability to assess comprehensive understanding. As a result, it tends to evaluate only partial and shallow comprehension rather than full understanding of the given contexts \cite{hsieh2024ruler, kuratov2024babilong}. While approaches such as \cite{li2024needlebench} have attempted to evaluate holistic understanding, they remain rudimentary and fail to provide a thorough analysis of the insights such methods can offer. In particular, these approaches often require models to simply identify ancestors, which only demands shallow reasoning and therefore falls short of providing a complete evaluation. In response, we propose a novel benchmark that evaluates holistic context comprehension and requires models not to miss any piece of evidence in the given context. Our benchmark and experimental results show that comprehensive understanding within a given context length is a high-stakes task that current models still struggle to handle.

\section{Conclusion}
In this work, we introduced \textsc{NeedleChain}, a context comprehension benchmark in which every part of the context contains query-relevant information. We designed \textsc{NeedleChain} so that the model cannot reach the correct answer if it misses even a single piece of evidence. Our experiments show that even when only 10 such evidence units exist, LLMs still struggle to fully comprehend the context. By constructing forward, backward, and mixed chain variants, we provided a comprehensive analysis of LLMs’ context understanding. Our results reveal that LLMs perform markedly worse on backward chains, where the reasoning order proceeds right-to-left, than on mixed-order contexts. We further proposed ROPE contraction, a simple yet compelling method to enhance context understanding abilities, achieving significant performance improvements on the \textsc{NeedleChain} benchmark. Our research indicates substantial room for improvement in the context comprehension capability of LLMs. Additionally, our analysis offers practical advice, suggesting that designing reasoning orders in a forward direction is beneficial when establishing long contexts. For future research, we aim to design benchmarks that encompass a wider range of domains.


\section{Limitation}
Due to resource constraints, we were unable to conduct extensive experiments with reasoning models. The excessive length of the input for the reasoning model made it impossible to perform benchmark evaluations under our limited resources. For instance, in a scenario with k=100, the QwQ model\cite{team2025qwq} required over 30 minutes to process a single query in our experimental setup. Given the need to process 200 queries for a single test, conducting a wide range of experiments with the inference model was impractical for us. Instead, we report the experimental results for $k$=50 in the Appendix~\ref{app:reasoning} and publicly release the data generation code to enable experiments with any higher $k$. We hope this will facilitate broader experimentation using our data in the future.

One limitation of our study is the exclusive use of needles requiring numerical calculation. However, through rigorously designed controlled experiments, we have clearly and robustly demonstrated our conclusions within the given setting. We plan to extend our benchmark paradigm to enhance its generalizability in future research.

\section*{Ethics Statement}
All names referenced in the dataset are fictional, as noted in Section 2.3, and are merely borrowed for illustrative purposes. We affirm that all salaries and names mentioned in the data bear no connection to real-world individuals. Additionally, any potentially referenced names contain no harmful content whatsoever. The models and methodologies employed adhere to community-accepted ethical practices and are not designed for harmful, malicious, or discriminatory purposes. We believe this work aligns with responsible AI research principles and does not introduce foreseeable risks of misuse.


\bibliography{custom}

\appendix

\section{Evaluation Details}
\label{app:evaluation_details}
We conducted all experiments using eight RTX A6000 GPUs. We performed decoding with a temperature setting of 0.6 and a top-p of 0.95, as these parameters represent the optimal prompt suggested in Qwen3\footnote{https://huggingface.co/Qwen/Qwen3-32B}. We utilized publicly available checkpoints from HuggingFace \cite{wolf-etal-2020-transformers} for all models and executed inference using the vllm \cite{kwon2023efficient} framework. Configuration for each model is as follows: Qwen2.5-32B \cite{qwen2.5} (\texttt{Qwen/Qwen2.5-32B-Instruct}), Qwen3-32B \cite{yang2025qwen3} (\texttt{Qwen/Qwen3-32B}), QwenLong-L1 \cite{yang2025qwen2} (\texttt{Tongyi-Zhiwen/QwenLong-L1-32B}), Llama3.3-70B \cite{grattafiori2024llama} (\texttt{meta-llama/Llama-3.3-70B-Instruct}), QwQ-32B \cite{team2025qwq} (\texttt{Qwen/QwQ-32B}), GPT-4o \cite{hurst2024gpt} (\texttt{gpt-4o-2024-08-06}).
The prompt employed for model evaluation is as follows:

\definecolor{judgecolor}{HTML}{F5F5F7}

\begin{table*}[h]
\centering

\footnotesize
\begin{tabular}{p\linewidth}
\toprule[1.5pt]
\rowcolor{judgecolor} \textbf{\#\# System Prompt} \\
\makecell[l]{
You are a financial assistant AI skilled in calculating wages and solving salary-related queries.\\
I will give you context with the facts about salary of several people. \\
You need to answer the question based only on the information from the facts. \\
Before you derive the final answer, provide me a brief explanation. \\
Output your final verdict by strictly following this format: '\#\# Answer: \$\{your\_answer\}'  \\
}\\
\rowcolor{judgecolor} \textbf{\#\# Input Template} \\
\makecell[l]{
There are \{num\_names\} workers in the office.\\
Their names are as follows: \{names\} \\\\
Salary for each worker is as follows:\\
\{chain\} \\\\
Now, respond to my question:\\
\{question\}
}\\
\bottomrule[1.5pt]

\end{tabular}
\caption{The default prompt for evaluating NeedleChain} \label{tb:prompt_eval}

\end{table*}

\section{Qualitative Analysis}
\label{app:qualitative}


\definecolor{legendcolor}{HTML}{F0E5D8}
\definecolor{legendcolor2}{HTML}{FFCDC9}
\definecolor{fontcolor1}{HTML}{00A8B5}
\definecolor{fontcolor2}{HTML}{774898}
\definecolor{fontcolor3}{HTML}{DE4383}
\definecolor{fontcolor4}{HTML}{F3AE4B}
\definecolor{fontcolor5}{HTML}{09009B}

\begin{table*}[h]
\centering

\footnotesize
\begin{tabular}{p{0.32\linewidth}|p{0.32\linewidth}|p{0.32\linewidth}}
\toprule[1.5pt]
\rowcolor{legendcolor} \multicolumn{3}{c}{\textbf{Chain Composition}} \\ 
\midrule[1.5pt]
\multicolumn{1}{r|}{\textbf{Common Context:}} & 
\multicolumn{2}{c}{
\makecell[c]{
There are 5 workers in the office.\\
Their names are as follows: 
\textcolor{fontcolor1}{Cairo}, 
\textcolor{fontcolor2}{Ramon}, 
\textcolor{fontcolor3}{Nala}, 
\textcolor{fontcolor4}{Kase}, 
\textcolor{fontcolor5}{Ryker}
}
}\\ \midrule
\makecell[l]{
\textbf{[Forward Chain]} \\
\textcolor{fontcolor1}{Cairo received \$1600 last week.} \\
\textcolor{fontcolor2}{Ramon earns twice as much as Cairo.} \\
\textcolor{fontcolor3}{Nala earns the same salary as Ramon.} \\
\textcolor{fontcolor4}{Kase earns the same salary as Nala.} \\
\textcolor{fontcolor5}{Ryker earns twice as much as Kase.}
} &
\makecell[l]{
\textbf{[Backward Chain]} \\
\textcolor{fontcolor5}{Ryker earns twice as much as Kase.} \\
\textcolor{fontcolor4}{Kase earns the same salary as Nala.} \\
\textcolor{fontcolor3}{Nala earns the same salary as Ramon.} \\
\textcolor{fontcolor2}{Ramon earns twice as much as Cairo.} \\
\textcolor{fontcolor1}{Cairo received \$1600 last week.}
} &
\makecell[l]{
\textbf{[Chaotic Chain]} \\
\textcolor{fontcolor2}{Ramon earns twice as much as Cairo.} \\
\textcolor{fontcolor5}{Ryker earns twice as much as Kase.} \\
\textcolor{fontcolor3}{Nala earns the same salary as Ramon.} \\
\textcolor{fontcolor1}{Cairo received \$1600 last week.} \\
\textcolor{fontcolor4}{Kase earns the same salary as Nala.}
} \\ \midrule
\multicolumn{1}{r|}{\textbf{Common Question:}} & 
\multicolumn{2}{c}{
\makecell[c]{
Now, respond to my question: \\
How much salary did \textcolor{fontcolor5}{Ryker} get?
}
}\\ \midrule[1.5pt]

\rowcolor{legendcolor}
\multicolumn{3}{c}{\textbf{Error Category}} \\\midrule[1.5pt]
\multicolumn{1}{c}{\textbf{Instruction not Followed}} & 
\multicolumn{1}{c}{\textbf{Needle Omission}} & 
\multicolumn{1}{c}{\textbf{Calculation Error}} \\ \midrule
...(explanation)... \#\# Answer: Amount adherent context depends suggestively Jayceon's half per explicit finalized quotas otherwise prescribed, rather undetermined under fact via customary series. &
...(three person missing)... \#\# Therefore, since Kyren equals Justin, which equals Salma, which is equal to Ronin: \$200. \newline\#\# Answer: \$200 &
...(explanation)... \#\# Answer: 1600 \$ \newline
(reference answer: 3200 \$)\\ \midrule

...(explanation)...  Thus, Miriam's salary in terms of (x) is: \#\# Answer: 4x &
...(five person missing) Therefore, Wrenleigh receives a salary of \$12.5.\newline\#\# Answer: \$12.50 & 
...(explanation)... \#\# Answer: 200 \$ \newline
(reference answer: 6400 \$)  \\ \midrule

...(explanation)... deductible as a function of referential salary "X" (Alora), as leaving Sol\'s clear deduction numerically unexpressed specific to definitively concrete values. Without Alora or alternate equitably lead measure detail.&
...(five person missing)... - Dayana earns twice as much as Wren. Therefore, Dayana earns \$25. \newline18. **Bridget and Dayana: - Bridget earns half as much as Dayana. Therefore, Bridget earns \$12.5.\newline\#\# Answer: \$12.5
&
...(explanation)... \#\# Answer: 2200 \$ \newline
(reference answer: 3200 \$)  \\
\bottomrule[1.5pt]

\end{tabular}
\caption{Data samples for each chain composition, along with qualitative error analysis.} \label{tb:samples}

\end{table*}

In this section, we extend the analysis from Section~\ref{sec:error_analysis} and provide specific examples for each error instance. Table~\ref{tb:samples} shows examples for each error category we defined, derived from GPT-4o outputs generated using the Mixed chain. Here, we also provide examples of the chain compositions we define. Each chain consists of a common context, a chain component, and a common question, and we treat the concatenation of these three elements as a single data point.

\textbf{Instruction not followed} also includes potential needle omission error. In this case, the model receives the relevant information but treats it as a free variable when producing the final answer. For convenience, we categorize such cases as instruction not followed because the model does not produce a deterministic final answer.

\textbf{Needle omission} denotes cases where the model reaches a final answer but does not mention any needle in its intermediate reasoning. We label these responses as incorrect \textbf{regardless of the correctness of the final answer}, in order to exclude lucky guesses. We justify this choice as follows: the explanation explicitly lays out the full reasoning process, so if the reasoning omits the needle, the model has not actually used the provided context.

Most \textbf{calculation errors} arise from mistakes involving powers of two. In a few rare cases, the model outputs a number that is not divisible by a power of two.

\section{Performance of Reasoning Model}
\label{app:reasoning}
Reasoning models demonstrate exceptional problem-solving for a variety of tasks. In this study, we analyze the performance of reasoning models on our benchmark. The experimental results are presented in Table~\ref{tb:reasoning}. Along with accuracy, we also report the length of the generated responses. Similar to existing models, reasoning models exhibit diminished performance in backward chain. Although the decline is less pronounced than that in traditional LLMs, we observe a clear trend of performance degradation with larger $k$. This indicates that achieving intact understanding remains a consistent challenge even for reasoning models.
\definecolor{metacolorab}{HTML}{F0F0F0}

\begin{table*}[h]

\centering
\resizebox{1.0\linewidth}{!}{
\footnotesize
\begin{tabular}{l|cccc|cccc|cccc}

\toprule[1.5pt]

\multicolumn{1}{c|}{\multirow{3}{*}{\textbf{Model}}} & 
\multicolumn{4}{c|}{\quad k=10 (Token Length: 0.1K) \quad} &
\multicolumn{4}{c|}{\quad k=20 (Token Length: 0.2K) \quad} &
\multicolumn{4}{c}{\quad k=50 (Token Length: 0.5K) \quad} \\

{} &
\multirow{2}{*}{\textbf{NS}} & \multicolumn{3}{c|}{\textbf{NeedleChain}} & 
\multirow{2}{*}{\textbf{NS}} & \multicolumn{3}{c|}{\textbf{NeedleChain}} & 
\multirow{2}{*}{\textbf{NS}} & \multicolumn{3}{c}{\textbf{NeedleChain}} \\

{} &
& \textbf{F} & \textbf{B} & \textbf{M} &
& \textbf{F} & \textbf{B} & \textbf{M} &
& \textbf{F} & \textbf{B} & \textbf{M} \\
\cmidrule(lr){1-1}\cmidrule(lr){2-2}\cmidrule(lr){3-3}\cmidrule(lr){4-4}\cmidrule(lr){5-5}
\cmidrule(lr){6-6}\cmidrule(lr){7-7}\cmidrule(lr){8-8}\cmidrule(lr){9-9}
\cmidrule(lr){10-10}\cmidrule(lr){11-11}\cmidrule(lr){12-12}\cmidrule(lr){13-13}

\textbf{Qwen2.5-32B} & \makecell[c]{100\\(36.41)} & \makecell[c]{100\\(283.76)} & \makecell[c]{91.5\\(284.09)} & \makecell[c]{93.5\\(306.14)} & \makecell[c]{100\\(33.27)} & \makecell[c]{97.0\\(500.44)} & \makecell[c]{53.0\\(505.64)} & \makecell[c]{76.0\\(568.13)} & \makecell[c]{99.5\\(31.98)} & \makecell[c]{84.5\\(1072.82)} & \makecell[c]{13.0\\(1175.93)} & \makecell[c]{25.5\\(1282.09)} \\ \midrule

\textbf{QwQ-32B} & \makecell[c]{100\\(300.95)} & \makecell[c]{100\\(618.09)} & \makecell[c]{96.5\\(1220.63)} & \makecell[c]{99.5\\(934.02)} & \makecell[c]{99.5\\(626.59)} & \makecell[c]{93.5\\(1036.64)} & \makecell[c]{76.0\\(2950.03)} & \makecell[c]{91.5\\(2196.24)} & \makecell[c]{100\\(875.13)} & \makecell[c]{86.5\\(2727.29)} & \makecell[c]{19.0\\(7938.16)} & \makecell[c]{62.5\\(6822.31)} \\ \midrule

\textbf{Qwen3-32B} & \makecell[c]{100\\(283.97)} & \makecell[c]{100\\(492.95)} & \makecell[c]{99.0\\(573.08)} & \makecell[c]{100\\(655.75)} & \makecell[c]{100\\(339.8)} & \makecell[c]{97.0\\(826.15)} & \makecell[c]{88.5\\(1031.43)} & \makecell[c]{96.5\\(1141.44)} & \makecell[c]{98.5\\(422.35)} & \makecell[c]{77.5\\(1810.63)} & \makecell[c]{24.0\\(3001.62)} & \makecell[c]{63.0\\(3368.24)} \\

\midrule[1.5pt]
\end{tabular}
}
\caption{Performance of reasoning LLMs on NeedleChain (\textbf{NIAH}: Needle Stack, \textbf{F}: Forward chain, \textbf{B}: Backward, \textbf{M}: Mixed Chain). We report both accuracy and the token length of the generated text.
}
\label{tb:reasoning}
\end{table*}

\section{LLM Usage Disclaim}
We utilized an AI writer solely for polishing the English expressions in the paper. Beyond this, no assistance was received in terms of ideation or other content development; the AI writer's role was strictly limited to correcting English expressions.


\end{document}